\documentclass[review]{elsarticle}
\usepackage{algorithm}
\usepackage{algorithmic}
\usepackage{lineno,hyperref}
\journal{Indian Institute of Technology - Kanpur}

\begin{document}

\begin{frontmatter}

\title{Soft is Safe: Human-Robot Interaction for Soft Robots}


\author{Rajashekhar V S \fnref{Corresponding author}}
\address{Research Scholar, Department of Design, Indian Institute of Technology - Kanpur, Uttar Pradesh, India}
\fntext[myfootnote]{Corresponding author}
\ead{raja23@iitk.ac.in}

\author{Gowdham Prabhakar}
\address{Assistant Professor, Department of Design, Indian Institute of Technology - Kanpur, Uttar Pradesh, India}
\ead{gowdhampg@iitk.ac.in}

\begin{abstract}
With the presence of robots increasing in the society, the need for interacting with robots is becoming necessary. The field of Human-Robot Interaction (HRI) has emerged important since more repetitive and tiresome jobs are being done by robots. In the recent times, the field of soft robotics has seen a boom in the field of research and commercialization. The Industry 5.0 focuses on human robot collaboration which also spurs the field of soft robotics. However the HRI for soft robotics is still in the nascent stage. In this work we review and then discuss how HRI is done for soft robots. We first discuss the control, design, materials and manufacturing of soft robots. This will provide an understanding of what is being interacted with. Then we discuss about the various input and output modalities that are used in HRI. The applications where the HRI for soft robots are found in the literature are discussed in detail. Then the limitations of HRI for soft robots and various research opportunities that exist in this field are discussed in detail. It is concluded that there is a huge scope for development for HRI for soft robots.     
\end{abstract}


\end{frontmatter}

%\linenumbers

\section{Introduction}
% talk about compliant, flexible and soft
The terms compliant, flexible and soft are often confusing when it applied to describe mechanisms. The compliant mechanisms rely on the intrinsic flexibility of materials for motion, while flexible mechanisms incorporate various flexible components such as cables and springs for movement \cite{howell2013handbook}. The soft mechanisms incorporate compliant and flexible elements to achieve adaptable and gentle motion \cite{chen2017soft}, often mimicking natural movements and interactions. A soft robot can be considered as a robotic system made from compliant materials that enables it to deform, bend, and adapt to its environment, offering flexibility and safety in human-robot interactions.

\subsection{Overview of Human-Robot Interaction}
The soft robots are safe and compliant when compared to rigid bodied robots during HRI. It was considered as a natural option for HRI due to their lower accidental impact forces and higher power density ratio \cite{jensen2022enabling}. The soft robots that are inspired by the living organisms can be used for performing safe HRI. The wearable electronics and soft robots emphasize for tactile and skin-friendly interfaces \cite{xiong2021functional}. There are ethical and philosophical aspects for shifting from rigid bodied robots to soft bodied robots which are discussed in \cite{hua2023rigid}. The progress in the field of human-robot collaboration till the year \textit{2018} can be found in \cite{ajoudani2018progress}. A review on HRI for soft robotics prior to \textit{2019} can be referred to in \cite{das2019review}. It discusses the bio-inspiration, modelling, actuation, control and applications in detail. A review of safe physical human robot interaction prior to \textit{2008} can be found in \cite{pervez2008safe}. An atlas of physical HRI (pHRI) discusses about the safety, mechanics and control issues, dependability and benchmarks set for their performance. It emphasises that the safety and dependability issues in pHRI still needs to be addressed \cite{de2008atlas}. These have been addressed in the recent years \cite{zacharaki2020safety}. The acceptance of machines and robots in the society relies on trust in the interaction that they have with humans. The trust was built on physical safety, operational understanding, and social training. The soft robots, which are adaptable in nature and use soft materials, enhance the safety and ease of operation \cite{van2021human}.

\begin{figure}
\begin{center}
\includegraphics[scale=0.40]{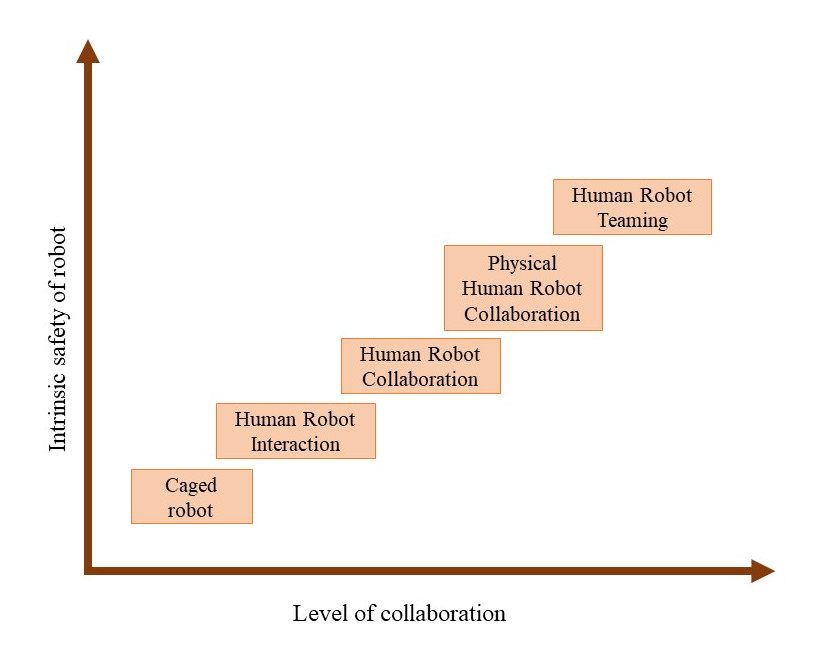}
\caption{The stages in human robot relationship \cite{hua2023rigid}}
\label{fig_hri}
\end{center}
\end{figure}

There are different stages in the levels of collaboration between the robots and humans \cite{hua2023rigid}. They are: (1) Caged robot (2) Human Robot Interaction (3) Human Robot Collaboration (4) Physical Human Robot Collaboration, and (5) Human Robot Teaming. This can be seen in Figure \ref{fig_hri}. In caged robots, there is no interaction between the human and the robot. The robot is placed inside a fenced structure where it does the intended task. In human robot interaction, the robot interacts and communicate with humans. This would be using input and output modalities. The human robot collaboration happens where the robot assists the humans to achieve shared goals. In physical human robot collaboration, the tasks are achieved cooperatively by the direct physical interaction between humans and robots. In human robot teaming, the humans and robots have a collaborative partnership to achieve shared goals through coordinated efforts. The level of collaboration and intrinsic safety of robots increase as the level increases. 

\subsection{Overview of Soft Robotics}
The soft robots have several advantages over the traditional rigid bodied robots \cite{sun2022softsar}. They have soft links, smoother movement control, highly flexible and deformable, and possess close to infinite degrees of freedom. These characteristics are absent in rigid bodied robots. A review on soft robots that are made of materials with elastic modulus ranging several kilopascals are intrinsically soft in nature. The manufacturing, sensing and control of these soft robots that were fluid driven and prior to the year \textit{2017} are discussed in \cite{polygerinos2017soft}. The developments in the field of flexible and stretchable electronics aid the better functioning and gives new direction to the research in soft robotics \cite{lu2014flexible}. The compliance in the soft hands that are attached as an end effector to a robotic arm takes care of the uncertainty in grasping and identifying the object of interest \cite{homberg2019robust}. A progress report that highlights materials and mechanics for soft robots were presented in \cite{majidi2019soft}. It highlights various actuating architectures such as fluid-elastomer structures, dielectric elastomers and thermal actuators.

The modelling of soft robots are important from the controls perspective. They are dealt with in detail in \cite{qin2023modeling}. A survey of the model based control of soft robots are presented in \cite{della2023model}. Also, the developments in the field of soft sensors and actuators aid the development of soft robots. The field of material science has matured in the past few decades and it contributes to the developments in the field of soft robotics \cite{gariya2021review}. The fabrication of soft robots are no more a challenge due to the developments in the field of manufacturing \cite{zhang2023progress}. The various soft robotic actuation modalities such as fluidic actuation, electrostatic actuation, electrochemical actuation, thermal actuation and magnetic actuation are explained in detail in \cite{yasa2023overview}. The various methods of sensing in soft robots are mentioned in \cite{hegde2023sensing}. With the advances in the field of modelling and control of soft robots, their fabrication, actuation and sensing, the field of soft robotics has been advancing on par with the rigid bodied robotics. 

\subsection{Human Robot Interaction applied to Soft Robotics}
The area of HRI for soft robotics focus on designing robots with soft materials and mechanisms to ensure safe and intuitive interaction with humans. With \textit{Industry 5.0} focusing on human robot collaboration, the HRI for soft robotics plays a very important role in creating a relationship between the humans and robots. There are several modalities that contribute for this interaction. Also the control, design and manufacturing of these soft robots are important from the user perspective since they interact with them. The advantage of HRI in soft robotics is the role of safety during interaction. Due to the soft nature of the robot, the injury caused to the humans can be minimized or absent. Unlike the HRI for rigid bodied robots, a broader spectrum of humans ranging from infants to elderly people can be benefited from the HRI for soft robots. Therefore this area in robotics has a wider scope for exploration and many contributions will arise in the upcoming years. 

\subsection{Aim and organization of the paper} 
The aim of the paper is to highlight and report the research works that have been done in the field of HRI for soft robots. It has been found that only in this millennium works have started to emerge in this area of robotics. In section \ref{subsec_screen}, the method used for screening and selection of articles that are reported in this work has been described. In Section \ref{sec_control_des_manu}, the work addresses the control, design, and manufacturing of soft robots, incorporating human-robot interaction. In Section \ref{sec_io}, the input and output modalities that are used for HRI of soft robots are presented. The Section \ref{sec_appli} describes the application of HRI in the field of soft robotics. This is followed by discussion in Section \ref{sec_discussion} and concluding remarks in Section \ref{sec_conclusion}. 

\subsection{Method for Screening and Selection of Articles}
\label{subsec_screen}
\begin{figure}
\begin{center}
\includegraphics[scale=0.40]{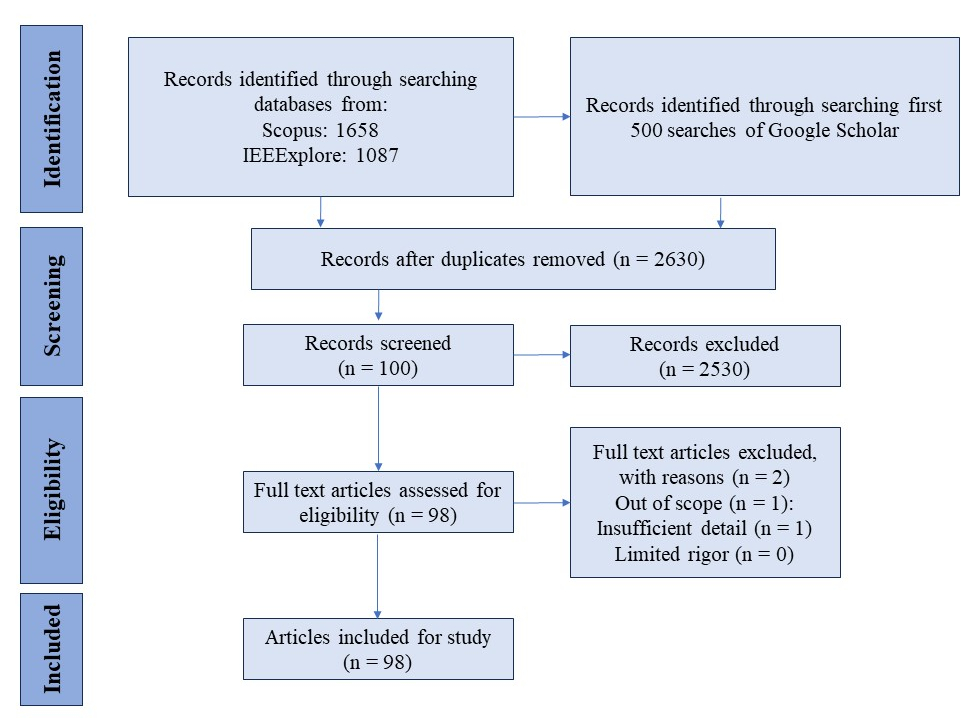}
\caption{The PRISMA flow diagram that shows the search results and screening}
\label{fig_prisma}
\end{center}
\end{figure}

\begin{figure}
\begin{center}
\includegraphics[scale=0.35]{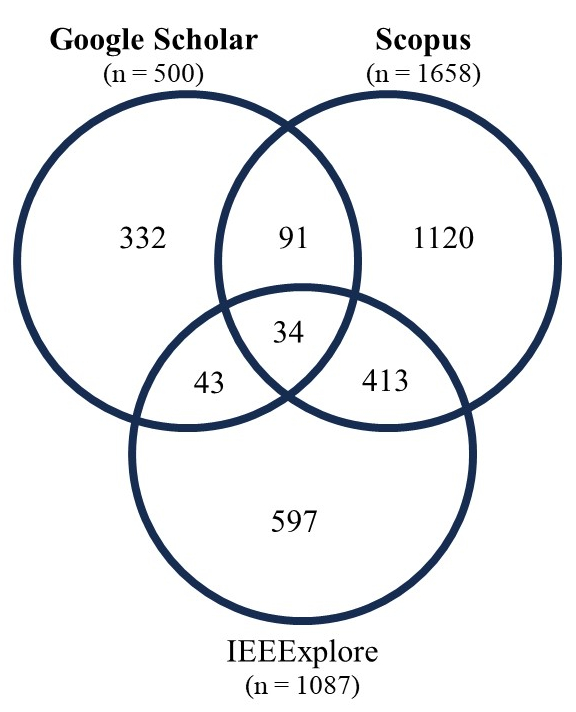}
\caption{The Venn diagram showing the duplicates in the three databases}
\label{fig_pie}
\end{center}
\end{figure}

The screening and selection of the articles were done based on the \textit{PRISMA 2020 statement} \cite{page2021prisma}. It can be seen in Figure \ref{fig_prisma} The first stage in this process was the identification stage where the \textit{Google Scholar}, \textit{Scopus} and \textit{IEEExplore} were chosen as the databases to search the articles. The key word \textit{"soft AND robot AND human AND robot AND interaction"} were used for searching in the databases. The details of the articles were downloaded in \textit{CSV} format from all the three databases. The titles were alone chosen in a separate \textit{CSV} file. In the second stage, the screening of the articles took place. A \textit{python} program, whose algorithm as presented in Appendix \ref{append_duplicate}, was used to identify the duplicate files in the three databases. A Venn diagram representing the results of finding the duplicate file identification has been shown in Figure \ref{fig_pie}. The records were screened based on their titles to determine their eligibility. The articles that dealt with human robot interaction for particularly soft robots were selected. There were about \textit{100} articles that were eligible as per the above mentioned category. In the third stage, the full text were accessed for all the articles and about \textit{98} articles were finally included for the study. Among the two articles that were excluded, one was due to out of scope and the other had insufficient details. It has to be noted that we focus our discussion in Sections \ref{sec_control_des_manu}, \ref{sec_io}, \ref{sec_appli} and \ref{sec_discussion} based on the \textit{98} articles that were screened using the above method.  

\section{Control, Design, Materials and Manufacturing of Soft Robots}
In this section, we discuss the various control methods that are used in the field of HRI for soft robots. Then we discuss the design of several soft robots on which the HRI has been performed. This is followed by the material selection and manufacturing procedure in brief for the soft robots, where the HRI has been done.  
\label{sec_control_des_manu}
\subsection{Control}
A survey on model based control of soft robots are presented in \cite{della2023model}. The works of \cite{ahmed2011compliance}, demonstrate the compliance control of a robotic manipulator for safe physical HRI. In \cite{stroppa2023shared}, shared control for teleoperation of a soft growing robotic manipulator was done. A master slave position control \cite{baiden2013human} was done for a 2 DoF exoskeleton robot. In \cite{zhang2019eeg}, a real-time control of soft robotic hand was performed. A model-based dynamic feedback control was performed for a planar soft robot in \cite{della2020model}. In the works of \cite{queisser2014active}, an active compliant control mode was used to interact with a pneumatic soft robot. In \cite{tang2021model}, model-based online learning and adaptive control for a human-wearable soft robot was done and presented. The impedance control of a hand-arm for HRI was done and presented in \cite{ficuciello2010modelling}. In \cite{ficuciello2012compliant}, active compliance control of soft fingers and force sensing for HRI was presented.  A model-based control algorithm for quasi-static regulation of motion and force in a soft robotic exoskeleton for hand assistance and rehabilitation was developed and presented in \cite{haghshenas2020quasi}. A cable-driven soft joint with torque-displacement modeling and a sliding mode controller demonstrated robustness in low-level torque control \cite{jarrett2017robust}. An adaptive quasi-static model-based control algorithm has been used to control a wearable (a soft robotic exo-digit) \cite{haghshenas2022adaptive}.

In the works of \cite{kim20153d}, pressure feedback controller was used to sense the contact and gently grasp the object. The soft robot module presented in \cite{zolfagharian2022ai}, was controlled using machine learning algorithms for safe pHRI. The impedance control of the soft robot named \textit{ALTER-EGO} was done in the works of \cite{lentini2019alter}. The same impedance control was done to the soft robot in \cite{de2008atlas} for a safe pHRI. The soft robot in \cite{kong2022bioinspired} was controlled using a customized deep neural network (DNN) algorithm. A hybrid controller for stiffness tuning and interaction control for shaping behaviours was aimed at ensuring safe interactions between the robot and the environment has been presented in \cite{yi2018customizable}. In the works of \cite{angelini2018decentralized}, the low-gain feedback action was merged with feed-forward action to control the soft robot that interacts with the environment. In \cite{li2020design}, the soft robot hand was controlled using a closed-loop PID control method for the flexion/extension angles of the robot hand.

The control method used to control the soft robot in \cite{jensen2022enabling} was the gain-based evolutionary model predictive control, specifically the model evolutionary gain-based predictive control (MEGa-PC). The research in \cite{hua2023rigid} highlights the importance of admittance and impedance control methods to manage interaction forces and compliance in HRI. The control method used to control the soft robot in \cite{armendariz2012manipulation} was an on-line bilateral teleoperation fuzzy controller. The impedance control was done in \cite{wolf2015soft} for the soft robot where, specifically joint level impedance behaviour which was implemented through a joint impedance controller with a back-stepping structure.

\subsection{Design}
The science of soft robot design has been explained in detail in \cite{stella2023science}. It looks in to how materials, morphology, control, and interactions happen with the environment. The bio-inspiration, computational, and human-driven design approaches are explored for better design of soft robots. Designing trajectory tracking controllers for soft robots, while maintaining their intrinsic compliance and achieving high performance was a challenging task \cite{angelini2018decentralized}. There were experiments that were conducted using a soft robot and force sensor \textit{(six-axis force/torque ATI mini 45 $^{TM}$)}. A low-gain feedback combined with iteratively learned feed-forward actions effectively preserves the softness in the robot. A force sensing at the finger tips of a soft robot was attached for grasping objects without slipping. Also the hand-arm control was done by exhibiting compliance when the force was applied to the finger tips \cite{ficuciello2012compliant}. A simulation was done by considering the soft pad and material properties. The results plotting the error, normal and tangential displacements were presented. A few of the soft robots in which the HRI takes place are presented as follows: 

\subsubsection{Soft Inflatable Joints for pHRI}
There are tendon wires which are pulled by linear actuators to operate the soft inflatable joints. A soft robot was built using multiple joints that were involved in pHRI. The physical interactions such as hugging tightly or lightly, thumping and handshake were performed on a soft humanoid robot that was built using the inflatable joints. The changes in air pressure during the physical interaction was observed and mapped which helped in identifying interactions \cite{niiyama2021blower}.   

\subsubsection{Interactive Gloves}
A soft robot was built using silicone and integrated to the virtual environment. The users interacted with the soft robot that was attached to the gloves that they were wearing. The gloves had as bending sensor to measure the bending of the fingers. The Unity software provides the virtual environment. The users wear gloves and interact with real objects, which adjusts the pressure and bending angle in the gloves according to the size of the virtual object \cite{eslami2023design}. It was planned to use this glove for wrist and finger rehabilitation.

\subsubsection{Soft Robotic Hand}
The potential of soft end effectors (SEEs) remains largely untapped, particularly in HRI \cite{bianchi2018touch}. A novel touch-based approach for soft end effectors with autonomous grasp sensory-motor capabilities by responding to objects passed to the robot by a human (human-to-robot handover) was presented. Upon contact, hand pose and closure are planned for grasping through arm motions executed with hand closure commands. These motions are generated from human wrist poses acquired as a human maneuvers the soft hand to grasp an object from a table. It was reported that the success rate was 86$\%$ for grasping various objects passed to the soft hand in different manners.

A soft robotic hand which has a fingertip haptic feedback can be used for safe HRI. The hand joint angles are collected using a data glove. The bending angles of the actuators in the soft robotic hand are measured using the flexion sensors. The robotic hand has pressure sensors in the fingertips to measure the forces. The operators have haptic feedback actuators present in their fingertips in order to display the contact forces. This set-up with a feedback control between the soft robotic glove and the user can be used for teleoperation \cite{li2020design}.  

The tendon driven soft robotic fingers were fabricated in \cite{ficuciello2010modelling}. It suggests a separate control law for the arm and fingers in order to better regularize the internal grasping forces. The interaction with the human and environment can be managed by using the arm. The force sensors at the finger tips regulate the contact forces for a stable grasp. 

\subsubsection{Soft Robots for Hugging}  
Designing robots that interact emotionally with humans, particularly children were studied in \cite{hedayati2019hugbot}. This could help children develop cognitive abilities by providing human like hugs that offers social and emotional support. The act of hugging was studied in humans by mounting pressure sensors on the shirt and then this data was planned to be transferred to a soft robot named \textit{HugBot} that can give human kind of hugs. 

A miniature humanoid robot featuring soft, air-filled modules at its joints, capable of detecting contact to prevent collisions that could harm both the robot and humans interacting with it was presented in \cite{kim2016study}. It was necessary to ensure the robot resilience and safety during these interactions. To assess the range of forces exerted during various types of hugs by children aged 4 to 10, a study was conducted involving 28 children. The pressure data was recorded as they softly and strongly hugged the pressure-sensing doll. The findings show that the maximum expected hugging force was \textit{2.623 psi} in the present set-up.

\subsubsection{Planar Soft Robot}
A experimental setup was made with a highly deformable soft robotic arm that had six segments with inflatable cavities which can be bidirectionally actuated. A model-based dynamic feedback control of the planar soft robot was done to enable interaction in unstructured environments. This was analytically analysed, simulated and experimentally verified \cite{della2020model}. 

\subsubsection{Soft Robotic Tentacle}
In the study that was conducted with a soft robotic tentacle, it was concluded that the appearance of the robot was an important aspect while designing soft robots \cite{jorgensen2018appeal} that involved human interaction. The tentacle can be controlled using infrared sensor that was present on the platform to which it has been fixed \cite{jorgensen2018interaction}. In the extended work by the same authors \cite{jorgensen2022soft}, the overall naturalness rating of the soft robot was statistically more than the rigid bodied robots. This was based on the user testing that was done using the above soft and rigid robotic tentacle.

\subsubsection{Modular Origami Soft Robot}
A soft origami robotic module that was reconfigurable and proprioceptive has been presented in \cite{huang2022modular}. A foldable self-inductance sensor was used to establish multimodal perception. Each module was capable of bending, and extending and contracting. Using three of these modules, an intelligent gripper was formed which grasped Yale-CMU-Berkeley (YCB) objects. With the addition of propellers and buoyancy chamber to it, an intelligent origami jellyfish robot was built using the origami modules. This robot was tested underwater where it grasped an object using the modules. 

\subsubsection{A Method for Analysis of pHRI}
An analysis method for predicting distributed loading across physical human-robot interaction (pHRI) interfaces was presented in \cite{yousaf2021method}. The method proposed improves the accuracy of predicting distributed interface loads, by considering compliance from human soft tissue and the attachment on the robot. The stiffness properties of a proxy upper arm was measured and used in the pHRI interface model. It was validated with the measurements from a sensorized upper arm cuff on the \textit{Harmony exoskeleton}. This confirms the effectiveness of the proposed method.

\subsubsection{Soft Robotic Exo-Digit}
An adaptive quasi-static control algorithm for a soft robotic exo-digit interacting with the human hand in continuous passive motion therapy was presented in \cite{haghshenas2022adaptive}. A discrete-time state-space representation for position control was done for the soft robot-human finger interaction by developing analytical models. The actuation pressure, which serves as the control input, was made to linearize the system and monitor the bending angle of the distal end. Experimental tests validate the controller's efficiency in responding to step inputs with disturbance suppression and accurately tracking the desired bending angle. An adaptive scheme fine-tunes control gains to accommodate parameter variations, guaranteeing stable and precise control throughout the operation.

\subsubsection{ALTER-EGO}
In the works of \cite{lentini2019alter}, a soft robot made of variable stiffness actuator was presented. The robot was capable of interacting with humans. The experiments show that the robot opens a door, interacts with children and carries heavy payload (10kg). The robot can manipulate objects and can move using the wheels. 

\subsubsection{SpineMan}
A novel spine-like manipulator (SpineMan) has been developed, featuring rigid elements made of polypropylene and soft elements consisting of polyvinyl alcohol (PVA) borax hydrogels enveloped in a silicone skin. The design was biologically inspired taking the vertebra as a base. The material selection and characterization for the manipulator was done and presented in detail \cite{runge2015spineman}. 

\subsubsection{Soft Growing Robot}
A human interface for teleoperating a soft growing robot in manipulation tasks using arm gestures was presented in \cite{stroppa2020human}. The participants completed pick-and-place tasks with high accuracy. The interface enabled intuitive control, with users following consistent strategies.

\subsection{Materials}
The authors in \cite{chen2021soft} built an actuator for a robotic finger using \textit{Dragon Skin 30$^{TM}$}. The filling of the finger was made of \textit{Ecoflex 00-30$^{TM}$}. An artificial skin \cite{teyssier2021human} utilizes various silicone elastomers to replicate the layers of human skin and incorporates an embedded electrode matrix for mutual capacitance sensing. A bladder used for making a soft robotic cuff that was used as a human robot interface was made of \textit{Shore 0030$^{TM}$} silicone rubber \cite{langlois2020investigating}. A soft robot that was built as an exo-digit for soft wearable HRI was made using RTV silicone rubber (RTV-4234-T4, Xiameter, Dow Corning) \cite{haghshenas2022adaptive}.

The soft robot in \cite{reitelshofer2015new}, was made using dielectric elastomer actuators (DEA), which are described as flexible capacitors with compliant electrodes enclosing a soft dielectric layer. In \cite{zolfagharian2022ai}, silicon rubber was the material chosen for the soft robotics module due to the elastic characteristics, moldability, durability, and resistance to mold and bacteria. The same silicone rubber was used in soft robot hand presented in \cite{zhang2019eeg} as the primary material. The soft robotic tentacle in \cite{jorgensen2018appeal} was cast in uncolored \textit{Ecoflex 0030} silicone with red wax. The soft robot in \cite{kong2022bioinspired} was made using a soft piezoresistive sensor based on a carbon black-coated polyurethane sponge. The \textit{CoboSkin} used PU foam and silicon rubber to produce a HRI skin \cite{pang2020coboskin}. The soft robot in \cite{eslami2023design} was made using silicon material, specifically \textit{Elastosil M4601}. The bending, abduction, and haptic feedback actuators in the soft robot hand \cite{li2020design} are 3D-printed using a soft material called \textit{NinjaFlex 85A} TPU.

The soft robot in \cite{kim2021design} was made using an inflatable body of low-density polyethylene (LDPE) and a highly sensitive and flexible strain sensor laminated onto it. In \cite{kozima2005designing}, the soft robot \textit{Keepon} was made of silicone rubber, giving it a simple, soft, and creature-like appearance. The soft robot in \cite{pang2018development} was made using a soft substrate consisting of polydimethylsiloxane (PDMS). In \cite{shembekar2021development}, the soft robot was made using a new type of material called permanent magnet elastomer (PME). The soft social robot in \cite{farhadi2022exploring} was made from \textit{Ecoflex 00-30} silicone. The soft growing robot in \cite{stroppa2020human} was made of a heat-sealable thermoplastic polyurethane fabric sheet. In \cite{jorgensen2018interaction}, silicone-based materials were used to make the soft robotic tentacle. The soft robot in \cite{armendariz2012manipulation} was made using soft-hemispherical fingertips made of polyurethane to grasp the object with improved contact. The soft inflatable robot was made using thermoplastic polyurethane (TPU) \cite{qi2014mechanical}, which has been an environmentally friendly material. For the soft robot in \cite{wang2024perceived} the robot was made using elastic materials such as \textit{Ecoflex 00-10} for the finger actuators and \textit{Dragon Skin 10 MEDIUM} for the palm. The soft robot in \cite{stroppa2023shared} was made using silicone material for its gripper and soft-growing manipulator. It can be inferred that soft polymers are widely used in manufacturing soft robots.

\subsection{Manufacturing}
The manufacturing of soft robots are done using several techniques which are discussed in this section. In certain scenarios, hybrid manufacturing has been done to build the robot. The manufacturing process used for the soft robot in \cite{pang2018development} involves inkjet printing and molding methods for the heterogeneous integration of components with different materials. The manufacturing process used in fabricating soft robots often include molding and casting \cite{homberg2019robust}.

\subsubsection{Injection Molding}
In this method, molded products are obtained by injecting raw materials that are heated by molds and then cooled and solidified. The manufacturing process used for the soft robot in \cite{lee2017durable} involves fabricating the piezoresistive elastomer using conductive nanofiller and silicone rubber, and then manufacturing it as a sensor using an injection molding process. 

\subsubsection{UV-laser-machining}
This manufacturing process involves using laser energy to break the bonding and form a new product. A UV-laser-machined stretchable multi-modal sensor network for soft robot interaction was presented in \cite{ham2022uv}. A sensor was built in Kirigami pattern using a flexible metalized film which was fabricated using UV laser metal ablation. This sensor can detect various external stimuli and was flexible in nature. This sensor network can be used for grasping warm objects or interacting with delicate surfaces.

\subsubsection{Aerosol-Jet-Printing}
This manufacturing process involves precise deposition of electronic Inks onto substrates. This process has been used as a novel manufacturing technique to generate the dielectric and electrode structures required for soft actuators and sensors based on dielectric elastomers. It allows precise deposition of very thin layers of printing material, ranging from hundreds of nanometers to several microns \cite{reitelshofer2015new}. 

\subsubsection{Casting and Molding}
This manufacturing process is identified by pouring molten material in the molds to form the required product. The manufacturing process in producing a soft robotic tentacle involves casting it in uncolored \textit{Ecoflex 0030} silicone with red wax, creating a three-chambered design that allows the tentacle to bend in various directions \cite{jorgensen2018appeal}. In \cite{farhadi2022exploring}, the manufacturing process involves casting the soft robot from \textit{Ecoflex 00-30} silicone using 3D printed molds. The manufacturing process of the soft robotic manipulation system in \cite{chen2021soft} involves mold-casting the soft arm into a bellow profile using \textit{Ecoflex 00-30}. The soft robotic arm and fingers in \cite{kim2016user} uses \textit{Ecoflex 00-30} as the material and manufactures them using a moulding process.

\subsubsection{Additive Manufacturing}
This manufacturing process is an incremental process where the product or the part is formed by a layer by layer technique. In the works of \cite{kim20153d}, a multi-material 3D printer was used to print the rigid and soft parts that were used in the robot. During the manufacturing of a soft robotic joint the manufacturing process involves 3D printing for fabricating the soft origami rotary actuators \cite{yi2018customizable}. The manufacturing process of the soft robot in \cite{eslami2023design} involves first developing a 3D model in CAD software, creating a mold using 3D printing, and then molding the robot using silicon. The fabrication process of the soft robot in \cite{li2020design} involves 3D printing the actuators with a \textit{LulzBot TAZ 6 Aerostruder} printer using NinjaFlex 85A TPU material. Silicone rubber was used to cover the surface of the bending actuators, and air tubes were glued to the actuators using silicone sealant. 

The manufacturing process used for creating the soft robot in \cite{alspach2015design} involves 3D printing, specifically PolyJet printing, which utilizes a gel-like material to support overhanging part geometry as each layer was printed. The manufacturing process soft robotic cuff in \cite{langlois2020investigating} involves molding the shell with liquid rubber in a 3D-printed cast and creating the inflatable bladder using a two-part mold. 

\subsubsection{Coating}
It is a process where a layer of material is deposited on the product or the part to prevent it from corrosion or scratches. The manufacturing process for the soft sensor used in the robot involves depositing carbon black (CB) on the polyurethane sponge to offer conductivity, making the sponge piezoresistive \cite{kong2022bioinspired}. 

\section{Input and Output Modalities}
\label{sec_io}
In the recent years research on control strategies for soft robots are being looked in to by researchers. In a work by \cite{stroppa2023shared}, the work done by humans and soft robots together on a shared basis was proved to be effectively. It concluded that constant haptic guidance would assist expert users efficiently and need-tailored guidance has been required for new users.   
\subsection{Input Modalities}
In this section we discuss the several input modalities that are used in the field of HRI for soft robots. These include compliance, gesture, brain-computer interface, touch, sensitive balloon sensor, robot skin, triboelectric nanogenerators and human face. They were found in the literature that we short-listed for this survey.

\subsubsection{Compliance}
It is the ability to bend for a given piece of material. The mechanical compliance plays an important role in physical HRI (pHRI). In a study which was done to  understand the behaviour of compliant link and compliant joint for pHRI \cite{she2020comparative}, concluded that the compliant links outperform the compliant joints while emphasising the safety in pHRI. 

A magneto rheological (MR) fluid based compliant actuator was used for safe pHRI. A two link planar robotic manipulator was built using the MR fluid based compliant actuator and safety analysis was experimentally done in static and dynamic environments \cite{ahmed2011compliance}.   

\subsubsection{Gestures}
The gesture is the ability to point and make a sign by an individual in order for the system or other human to understand some information \cite{bartneck2020human}. A soft module was fabricated where slapping, squeezing, and tickling were used as input gestures to the modules \cite{zolfagharian2022ai}. The module was made using silicone and two polyvinylidene fluoride (PVDF) sensors were attached to it. The working of the system is as follows: (1) The input gesture is read from the user (2) The vibration is collected by the PVDF sensor (3) The data is processed and filtered (4) The features are extracted (5) The classification is done using a machine learning algorithm (6) The recognized gesture is verbally communicated. The input modality for the soft growing robot \cite{stroppa2020human} were the gestures of the operator tracked by a motion capture system, which are mapped to the kinematics of the robot for teleoperation.

\subsubsection{Brain-computer interface (BCI)}
This interface reads signal from the human brain and assigns certain action for a particular signal \cite{bartneck2020human}. By combining electrooculography (EOG), electroencephalography (EEG), and electromyogram (EMG), a new multimodal human-machine interface (mHMI) was developed. The system had a soft robotic hand that was pneumatically operated. By a combination of hand gestures and eye movements in the EOG, EEG and EMG mode, it was used as a set-up to improve the motor function of stroke survivors \cite{zhang2019eeg}. 

\subsubsection{Touch}
The ability to interact by using mostly fingers (by contact) to convey information is known as touch \cite{bartneck2020human}. A tactile sensor was built for HRI using soft piezoresistive sensor (carbon black-coated polyurethane sponge). A deep neural network was used to train so that the sensor can recognize the five touch modalities that includes 1, 2, and 3 point press, pat and fist press. The sensor was then tested on a medical assistive robot named \textit{"CureBot"} \cite{kong2022bioinspired}. The involvement of affective touch in HRI with a haptic creature was studied in \cite{yohanan2012role}. A touch dictionary was created with gesture label and gesture definition. A user study was conducted where the participants had to choose the a gesture and their emotions were mapped in a table. This mapping will help the designers to build a better social robot with a easy to interact HRI.

The safety during tactile engagements was done by enabling high-resolution sensory readings through depth camera imaging \cite{huang2020high}. The validated data-driven model translates point cloud data into contact forces. It was demonstrated in real-world applications where the robot reacts to tactile input and provides dynamic assistance to human upper limbs. An innovative method for creating compliant artificial skin sensors for robots that mimic the mechanical characteristics of human skin and accurately detect tactile stimuli has been presented in \cite{teyssier2021human}. The sensor was attached to the \textit{Nao} humanoid robot and \textit{xArm 6} robotic arm for experimental verification.

A physical haptic feedback mechanism to facilitate intuitive transfer of human adaptive impedance to robots by utilizing electromyography (EMG) signals for limb impedance extraction was presented in \cite{yang2017interface}. By integrating spectral collaborative representation-based classifications and fast smooth envelop algorithms along with direct trajectory transfer, the interface enhances skill transfer in pHRI tasks. A highly strain-tolerant and linearly sensitive soft piezoresistive pressure sensor has been proposed for wearable HRI \cite{kim2024mechanically}. This sensor was mounted on a gripper of a robotic arm which converted the applied pressure to the electrical signals. The input modality for the soft robot in \cite{shembekar2021development} uses tactile sensing, where the skin can detect forces and pressures. 

\subsubsection{Sensitive Balloon Sensor}
The sensitive balloon sensor is used as an input modality where it is soft in nature. The tactile sensors play an important role in enhancing HRI. In the works of \cite{kim2021design}, a novel balloon sensor, composed of low-density polyethylene and a flexible strain sensor, was proposed. It was used to detect contact and prevent collisions in rigid bodied robots. The sensor was lesser in weight (2 g).

A novel soft inflatable arm designed for telepresence robots that enables remote interaction while mimicking human arm movements was presented in \cite{qi2014mechanical}. The proposed arm was lightweight (approximately 50 grams) and was driven by three small cables installed at the shoulder and elbow joints. It can be extended to have multiple joints by using more cables. The structure was inflatable and can have safe direct human contact without external sensors. 
 
\subsubsection{Robot Skin}
Robot skin serves as the interface between collaborative robots (cobots) and their external environments, facilitating timely responses to collisions during tasks to prevent damage. In industries, there are situations where the robots need to collaborate and work with humans to perform a task. A flexible tactile sensor that was attached to the body of the robot can sense obstacles. A flexible tactile sensor array was created so that it was spread over an area where the HRI happens. In a study by \cite{pang2018development}, these sensors were mounted on the YuMi robot and experiments were conducted. The results showed that the robot skin (array of tactile sensors) provided natural and secure HRI.

A durable and repairable soft tactile skin for robots using deformable piezoresistive elastomer and electrodes were proposed \cite{lee2017durable}. The electrodes were placed in the sensor boundary. The silicone elastomer sensing material allows easy repair of damaged area. The tactile skin allows sensing of pressure change using the resistivity reconstruction method. This tactile sensor allows pHRI and presence in assistive robots. An investigation of the impact of active and passive protective soft skins on collision forces in HRI was presented in \cite{svarny2022effect}. This work focuses on the power and force limiting regime as described in ISO/TS 15066 which was "Robots and robotic devices – Collaborative robots" by the International Organization for Standardization (ISO). The study deals with comparison of impact forces, their durations, and impulses in collisions with and without protective skins on different robotic manipulators. The results show that soft skins when attached to the robot can reduce peak impact forces and prolong collision durations which enhances safety in collaborative applications.

A collaborative robot skin (CoboSkin) that can sense and cushion collisions was proposed for effective HRI \cite{heng2021fluid}. It was done using a fluid-driven soft robot skin which integrates sponge force sensors and pneumatic actuators to achieve this. It was capable of offering tunable stiffness and sensitivity, thereby reducing collision force peaks. A more detailed analysis and implementation of the skin on the \textit{ABB YuMi} dual-arm cobot for safer HRI was presented in \cite{pang2020coboskin}. A 3D printed soft skin module with an airtight cavity for contact sensing using pressure sensor and gentle grasping was presented in \cite{kim20153d}. The pressure sensor also helps in interacting gently with soft objects. The skin module reduces impact forces during collisions, which enables safe HRI. This was experimentally verified by impacting masses ranging from 50g to 500g on the skin module. A soft upper body of a humanoid was built using the same soft skin module \cite{alspach2015design}. It had the ability to interact with the humans especially children.  

In the works of \cite{shembekar2021development}, a permanent magnet elastomer (PME) was developed. It was created by blending neodymium particles into a polymer base and magnetized up to 6 Tesla for strength and anisotropy, where the traditional hard permanent magnets were replaced in hall effect-based tactile sensors. To cover curved surfaces of robotic arms and other similar robots, a curved tactile sensor was designed and presented \cite{mukai2006development}. The curved tactile sensor was built using an array of pressure sensors covered with elastic membrane. The efficacy of the sensor was tested by mounting it on a robotic arm.

The skin-inspired triple tactile (SITT) sensors (SITT) are integrated into robotic fingers to facilitate precise bimanual grasping, featuring skin-inspired multilayer microstructures housing interdigital electrode, flexible force, and temperature sensors \cite{zhao2023skin}. These sensors enable simultaneous or independent measurement of a material's dielectric property, tactile force, and temperature. This sensor can be used for sensitive object handling, adaptive manipulation, and interactive robotics applications. A soft robot skin comprising of force-sensing units and a porous substrate with a serpentine structure was developed and presented in \cite{ye2022soft}. This design enables the skin to conform to the cobot surfaces with minimal stress. It provides high sensitivity and fast response for real-time collision detection. This was validated through finite element analysis and experimental verification on a cobot arm named \textit{YuMi}. 

\subsubsection{Triboelectric Nanogenerators}
The triboelectric nanogenerators (TENG) can be used as sensors in the field of robotics and  human–machine interfacing \cite{wang2020triboelectric}. A TENG-based HMI patch which was worn on the fingers that when slides results in a voltage signal. Based on the signal, a remote operation of a soft manipulator was carried out \cite{chen2021soft}.

A flexible bimodal smart skin (FBSS) was made using TENG \cite{liu2022touchless}. The FBSS was capable of tactile and touchless stimuli. The humans can teach the soft robotic arm with bare hand-eye coordination. The robot can then replay the commands that the humans taught it to do. This non-programmable teaching method can be used in several other domains of soft HRI.

\subsubsection{Human Face}
The human facial expressions play a vital role in everyday communication, conveying diverse information that our brains quickly interpret to facilitate interaction \cite{bartneck2020human}. A study investigates biometric and soft-biometric traits derived from facial information, using a single feature set that encompasses identity, gender, and facial behaviour \cite{riaz2011spatio}. Utilizing spatio-temporal multifeatures (STMF) extracted from image sequences via a 3D face model, this approach offers robustness against changes in head poses and facial expressions. Experimental evaluations conducted in lab environments captured images and three benchmark databases across varying poses and expressions provide insightful comparisons with alternative methodologies.

\subsection{Output Modalities}
In the works of \cite{amaya2021evaluation}, a comparison was made with two methods of control for a soft robotic manipulator. They are direct method, where Geomagic Touch$^{TM}$ was used, and indirect method, where HTC Vive controllers and head-mounted display was used. A user study was conducted where it was concluded that the indirect method had few errors and more usable to the participants.

\subsubsection{Actuators}
Soft actuators, such as pneumatic artificial muscles, are commonly used in soft robots \cite{hua2023rigid}. The field of actuators for soft robots have also developed to a greater extent which leads to advancements in this area. Some of the actuation methods include pneumatic actuation, vacuum actuation, cable-driven actuation, shape memory alloy actuation, electroactive polymer actuation and electro-adhesive actuation \cite{zaidi2021actuation}. The work \cite{hua2023rigid} suggests the usage of soft actuators for HRI since it has improved safety and minimized risk of harm to humans. The soft robot in \cite{shembekar2021development} uses a PME infused skin for tactile sensing, which acts as both the sensor and actuator. The cable-driven soft exoskeleton uses a series elastic actuator (SEA) as the actuator \cite{jarrett2017robust}.

\paragraph{Pneumatic Artificial Muscles}
The soft robot hand uses pneumatic artificial muscles (PAMs) as actuators for controlling the hand movements \cite{zhang2019eeg}. The soft robot in \cite{kim2021design} uses pneumatic artificial muscles and inflatable sleeves as actuators. The soft robot presented in \cite{singh2019design} uses a pneumatic artificial muscle actuator for HRI applications. The soft robot in \cite{amaya2021evaluation} was made of flexible material, and the soft bending modules are powered by pneumatic muscle actuators (PMAs). 

\paragraph{Variable Stiffness Actuators}
The robots such as \textit{DLR Hand Arm} are actuated using variable stiffness actuators. The actuator and the link are coupled by the mechanical spring. The stiffness of the spring was changed by the action of the actuator. Based on this stiffness, the interaction of the robot with the humans take place \cite{wolf2015soft}. Also a robot named \textit{ALTER-EGO} uses this variable stiffness actuator in the arms to interact with humans \cite{lentini2019alter}. A robotic skin named \textit{CoboSkin} exhibits variable stiffness  \cite{pang2020coboskin} by adjusting air pressure in the inflatable units that it was made.

\paragraph{Pneumatic Actuators}
The bionic soft robots, particularly pneumatically operated robots, offers flexibility and safety for physical interactions \cite{queisser2014active}. A hybrid approach integrating classical and learning elements proposes an interactive control mode for elastic bionic robots. This approach combines low-gain feedback control with feed-forward control using a simplified model of inverse dynamics. This type of control schemes can benefit other soft robots and pave way for broader applications in manipulation tasks. Also in \cite{baiden2013human}, pneumatic actuators are used as soft actuators in a rehabilitation device. The soft robotic tentacle in \cite{jorgensen2018appeal} uses a pneumatic actuator controlled by low noise pumps. The soft robot in \cite{eslami2023design} uses pneumatic actuators for its movement, which are controlled based on the commands received from the virtual reality environment. In \cite{li2020design}, pneumatic actuators are used in soft robots, where the air chamber was filled with pressurized air to enable bending, and returning to the original state due to material elasticity, allowing the soft robot hand to perform complex tasks. 

The pneumatic actuators are used in the soft robot for actuation which are controlled by electrical pumps and solenoid valves \cite{farhadi2022exploring}. The actuator used in soft robots of \cite{heng2021fluid} was a pneumatic actuating cell made of silicon rubber for the internal part and inelastic nylon textile for the external part. The actuating cell exhibits adjustable stiffness similar to natural muscle. The soft robot tentacle in \cite{jorgensen2018interaction} was controlled using a pneumatic actuation system with low-noise pumps.

\paragraph{Soft Origami Rotary Actuators (SoRAs)}
The design, modelling and fabrication of the SoRA has been presented in \cite{yi2018customizable}. A method to obtain a customized force was proposed using quantitative methods. A stiffness tuning and interaction control hybrid controller was proposed for safe interactions. The static and controller test results were reported in detail.

\paragraph{Soft Bidirectional Bending Actuator (SBBA)}
The SBBA was made up of soft pneumatic actuator. This actuator was pneumatically operated with pneumatic tube running through the body of the robot. This actuator was bidirectional in nature and can operate at low pressure. It has a wider workspace due to the absence of inextensible central layer. The kinematic model of this actuator was simple to implement \cite{singh2019design}. 

\paragraph{Fluid-based Actuator}
The conventional robot joints possess high-ratio gear trains, which can potentially inflict discomfort or pain upon human contact. A soft manipulator made of a smart flexible joint utilizing an ER fluid was presented in \cite{nakamura2003development}. Additionally, a torque controller was devised based on human pain tolerance, ensuring gentle interaction. The soft robot that was compliant in nature was made using magneto-rheological (MR) fluid based actuator \cite{ahmed2010semi}. A novel soft fluidic actuators featuring rotary elastic chambers (RECs), capable of producing rotational motion without additional transmission elements was presented in \cite{ivlev2009soft}. These actuators possess inherent compliance, high back-driveability, and a lightweight design, facilitating safe physical interaction with humans despite generating significant torque. With their compact design, REC actuators offer improved integration into complex kinematic structures compared to conventional linear-type fluidic muscles. The compliance of these actuators can be adjusted by regulating pressure in antagonistic chambers, making them suitable for transitioning from continuous passive motion to active assistive behavior during therapy sessions. The soft robotic manipulator \cite{ahmed2011compliance} was made using magneto-rheological (MR) fluid-based compliant actuators. The soft robot in \cite{pang2018development} uses fluidic elastomer actuators for safe natural safe HRI. 

\paragraph{Fiber Reinforced Elastomeric Enclosures (FREEs) Actuator}
The FREEs was a soft actuator that can be used both individually and in modules \cite{buffinton2020investigating}. It was made of thin latex tubes with helically wound cotton fibers. The findings indicate that Ogden material model, rather than neo-Hookean, better describes the behavior of elastomer materials under significant deformations when modeled with 1D truss elements for fibers. The material properties of the elastomer significantly impact FREE extension, expansion, and rotation, with strains exceeding 25$\%$, while variations in fiber stiffness had minimal effects on deformation.

\paragraph{Tendon Driven Actuators}
The tendon-driven soft wearable robots provide simplicity, compactness, and safety, given their direct interaction with the human body. Unlike rigid robots, wire pre-tension removal was essential at the end-effector due to the inherent nature of soft robots. A novel linear actuator, termed as slider-tendon linear actuator was designed to achieve stable and compact actuation without pre-tension. Since it uses tendons, this actuator was smaller in size when compared to ball screw or lead screw based linear actuators. The proposed actuator was used to actuate a wearable soft robot \cite{kim2021slider}. Also in the works of \cite{niiyama2021blower}, tendon wires pulled by linear actuators are used as the actuator in the soft robot to drive the inflatable joints. The soft robotic finger use actuators \cite{ficuciello2010modelling} that are tendon driven. The soft robot in \cite{ford2023tactile} uses tendon-driven actuators that actuate a network of tendons routed through the robotic hand.

\subsubsection{Physical Warmth}
The physical warmth from the robot has induced the feeling of trust and friendship in HRI. In the study \cite{nie2012can}, it was concluded that the act of handholding and physical temperature has an effect on human perception of robots. By wearing a custom made outfit, the \textit{PR2} robot was made to interact with humans in hard-cold, soft-cold and soft-warm experiment conditions \cite{block2019softness}. It was proposed to make the robots closer to humans and engage in social-physical HRI.

\subsubsection{Sound}
There were two forms of soft robots namely \textit{SONO} and \textit{Tentacles} that reacted to three different sound designs by moving in a pre-programmed path. The participants where asked to perceive them and then respond how they felt. Although the sound design was not statistically significant on participant perception in the social attribute of the robots, the sound design interpretation depend on robot type \cite{jorgensen2021sounds}.

\subsubsection{Tactile}
The ethical tactile considerations for designing soft robots for HRI was discussed in \cite{arnold2017tactile}. It highlights the need to ensure user trust and benefit understanding in HRI. It also discusses the bonding that arises between the robot and human due to tactile interface. The application of the soft robot was in human-robot interaction tasks, specifically in tasks requiring tactile sensing capabilities for detecting external stimuli and interactions \cite{kong2022bioinspired}. 

\subsubsection{Expressions}
A non-humanoid soft robot powered by pneumatic actuators was developed for fostering social interaction with humans (HRI) \cite{farhadi2022exploring}. The robot utilizes motion and gestures, such as tilting, expanding, and breathing-like movements for communication purposes. The eight movement pre-sets were manually coded to depict specific actions, internal states/emotions, or motion patterns: greeting, avoidance, breathlessness, joyfulness, alarm, jellyfish-like movements, fear, and sighing. An on-line survey involving 59 participants was conducted to assess their perception of each pre-set behaviour through word selection. The semantic analysis of participant word choices indicate successful conveyance of intended meanings for most pre-sets. The text mining techniques applied to general comments revealed perceptions of the robot resembling an animal-like sea creature or human body parts (e.g., lungs, belly, or heart). These findings show that the soft robotic movement as a means of communication with users. It can find applications in affective interfaces and social robotics design.

\subsubsection{Handshakiness}
The engagement of a human in a handshake with the robot enhances the perception of warmth, animacy and likeability \cite{avelino2018power}. A two benchmark experiments measuring contact locations in human-human/human-robot handshakes and pressure distribution with a sensorized palm was performed in \cite{knoop2017handshakiness}. These benchmarks allowed to measure various properties of human handshaking, including hand contact area, contact pressure distribution, and grasping force. This work serves as a reference point for comparing robot handshake and human handshaking dynamics.

\subsubsection{Collision Safety}
The safety guidelines that exist for rigid bodied robots in HRI was not applicable for soft robots. A framework was proposed in \cite{wang2024perceived} to perceive safety in human–soft robot interaction. An user study was conducted using quantitative and qualitative methods. A soft robotic finger was built and 15 interactive motions were done with the users. A few of the interactions include tapping, stroking, poking, grasping and pinching. The perceived safety scores were derived at the end of the user study. Also, the gaze data of the participants were recorded during the experiment. Then the gaze heatmaps were drawn and it showed that the participants looked at the soft robotic hand most of the time. 

Tasks involving human-robot interaction (HRI) necessitate robots capable of safely sharing workspace and exhibiting adaptable compliant behavior to ensure collision safety and maintain precise positioning. Traditionally, robot compliance control relies on active compliance control using sensor data, while passive devices offer controllable compliance motion albeit with mechanical complexity. A novel approach utilizing a semi-active compliant actuation mechanism employing magneto-rheological (MR) fluid-based actuators, introducing reconfigurable compliance characteristics into robot joints was presented in \cite{ahmed2010semi}. This method ensures high intrinsic safety through fluid mechanics and offers a simpler interaction control strategy compared to existing approaches. The performance of robot manipulator was experimentally validated, ensuring safe HRI by addressing static collision scenarios.

\section{Applications}
\label{sec_appli}

\subsection{Manipulation}
The \textit{ALTER-EGO} soft robot in \cite{lentini2019alter} has soft underactuated hands that can be used for manipulation task. The application of the soft robot in \cite{pang2018development} was to enhance the perception of external force by collaborative robots, enabling safe and natural human-robot interaction in smart factories. It can be noted that in smart factories the HRI has been a common activity.

\subsubsection{Soft-fingertips for Safe pHRI}
A soft fingertips for interacting with rigid objects through an online bilateral teleoperation fuzzy controller by eliminating time delay was presented in \cite{armendariz2012manipulation}. The fuzzy inference engine continuously adjusts the force feedback gain to enhance awareness of physical interaction, facilitating successful grasping. Unlike rigid fingertips, which assume an infinitesimally small contact point, this approach enables safe and intuitive interaction, as demonstrated in an experimental study involving 19 participants. The results indicate successful feedback of slave contact force to the human user, with subjects consistently reporting comfort and ease of manipulation using an experimental two-hand prototype.

\subsubsection{Tactile-Driven Gentle Grasping}
A control strategy for delicate grasping using a Pisa/IIT anthropomorphic \textit{SoftHand} equipped with a compact \textit{TacTip} optical tactile sensor on each of the five fingertips was presented in \cite{ford2023tactile}. These tactile sensors record data on grasping and finger-object interactions that facilitates the development of hardware advancements to enable rapid real-time grasp control through asynchronous sensor data acquisition and processing. A gentle and stable grasping of 43 objects with diverse geometries and stiffness profiles was successfully done in a human-to-robot handover scenario.

\subsubsection{Soft growing robot}
The soft growing manipulator can move through cluttered environments without being restricted by body parts that may collide with obstacles, making it beneficial for manipulation tasks \cite{stroppa2020human}. The application of the soft robot in \cite{stroppa2023shared} includes tasks such as pick-and-place operations, where the robot can grow, retract, steer, and manipulate objects in three dimensions.

\subsection{Medical and Healthcare}
The soft robots can be used in the field of medicine.  It requires a synergy between doctors, robotics devices, and patients for successful outcomes. The soft robots in the form of sensorized robotic patients can be used by doctors for practice and training \cite{scimeca2020human}. 

The safety and acceptability are of greater concerns in healthcare robotics, where traditional rigid robot arms pose potential risks to users upon impact. The soft robotic arms offer a safer alternative, yet their acceptability remains underexplored. The study in                                           \cite{kim2016user} investigates people reactions to soft robotic arms and fingers compared to conventional rigid forms and human counterparts, aiming to understand their perceived usefulness for healthcare tasks. The blindfolded participants were engaged in tactile tasks involving different arm and finger types. The results revealed that soft arms were deemed more human-like but fragile compared to hard arms. Additionally, overly soft fingers were perceived as creepy and less reliable.

\subsubsection{Companionship}
A soft robot named \textit{Probo} can be used as a test-bed for HRI. It has eyes, eyelids, eyebrows, ears, mouth, neck and trunk. They are actuated independently and can be used in HRI studies, especially for hospitalized children. This robot can inform and comfort the children. This robot was briefly presented in \cite{goris2009probo}. The application of the soft robot in \cite{alspach2015design} was for physical human-robot interaction, particularly in scenarios where the robot needs to interact with children in a safe and playful manner. The soft skin modules provide protection, contact sensing, and compliance during interactions. The application of the soft robot in \cite{farhadi2022exploring} was for social HRI, where the robot communicates with a human interaction partner through bodily motion and posture, resembling breathing movements. The application of the soft robot, HugBot \cite{hedayati2019hugbot}, was to provide human-like hugs to children in order to teach social skills through emotional interactions like hugs. The application of the soft robotic arm and finger in \cite{kim2016user} was intended for healthcare tasks, such as assisting disabled individuals in performing everyday activities like brushing teeth and picking up objects from a table.

\subsubsection{Strapping Pressure of Soft Robotic Cuff}
The significance of pressure distribution during the attachment of an exoskeleton to the human body was studied in \cite{langlois2020investigating}. A novel adaptive interface aimed at enhancing the connection between the exoskeleton and the individual was presented. A 7-DoF collaborative robot named \textit{Franka} was connected to the interface which the human was wearing. The participant was allowed to trace a trajectory where the pressure distribution in the interface was recorded. This helped in understanding the connection stiffness between the robot and the human.

\subsubsection{Motor Rehabilitation}
The application of the soft robot hand can be in motor rehabilitation therapies and assistive technology, specifically for hand rehabilitation and task-specific training \cite{zhang2019eeg}.  

\subsubsection{Stroke Rehabilitation}
A soft robotic glove for stroke rehabilitation was built using a soft-elastic composite actuator (SECA). A model-based online learning and adaptive control algorithm was proposed for the glove \cite{tang2021model}. 

\subsubsection{Autism Spectrum Disorder Therapy}
The children with ASD were promoted to have a pHRI with a social robotic platform that mas made of flexible elements. A user study was conducted with 35 children diagnosed with autism. The results show that the children had higher emotional response and tactile interaction with the robot when they were allowed to interact with the robot physically. This was statistically shown in\cite{pinto2022physical} post the study that was conducted between the robot, the child, and the therapist. 

\subsubsection{Rehabilitation}
A novel approach to continuously control the robot motion and providing targeted torque for patient rehabilitation by utilizing surface electromyography (sEMG) and a human neuro-musculo-skeletal model was presented in \cite{shi2014design}. This method translates the sEMG neural command signals into kinematic variables for controlling the rehabilitation robot which enables it to guide patients along proper trajectories and provide necessary force support. The application of the soft robot in \cite{eslami2023design} can be in rehabilitation applications, particularly in virtual reality settings, where it can interact with objects of varying sizes and conform to the angles of human fingers.

\subsubsection{Pervasive Developmental Disorders}
A robot named \textit{Keepon}, which was a small, 12cm tall, silicone rubber robot resembling a yellow snowman was aimed at studying human social development and potential interventions for developmental disorders \cite{kozima2005designing}. It involved a longitudinal observation spanning over a year involving more than 500 child-sessions. It took place in an unconstrained playroom where the interactions between children with developmental disorders and the robot happened. The findings indicate that children understanding and interaction with the robot evolved over time.

\subsubsection{Orthoses}
A set of compliant pneumatic soft-actuators are attached directly to the legs of the hemiplegia patient. The simplified control concept involves trajectory calculation performed by the motion controller, torque or pressure control within the soft-actuator subsystem, and the human wearing both an active and a passive orthosis \cite{baiden2013human}. 

\subsubsection{Lower Limb Rehabilitation}
The  assessing of interaction force between users and lower-limb gait rehabilitation exoskeletons was proposed in place of conventional load cells for distributed measurement of normal interaction pressure across the entire contact area. This method involves the insertion of a soft silicone tactile sensor between the limb and standard connection cuffs which allows precise assessment of pressure distribution throughout the interaction. This approach maintains comfort for the user and was adaptable to cuffs of varying shapes and sizes, and offers cost-effective manufacturing \cite{de2010soft}.

\subsubsection{Physical therapy}
The application of the soft robot presented in \cite{haghshenas2022adaptive} can be used for continuous passive motion physical therapy, specifically for applying slow repetitive motions to the passive user hand.

\section{Discussion: Limitations and Research Opportunities}
\label{sec_discussion}

\subsection{Limitations of HRI for Soft Robots}
The HRI for soft robots lack stability and control. Although this has been addressed in the works of \cite{della2023model}, there are a lot of open challenges that needs to be addressed. In the safety aspects, the pneumatic actuators pose concerns due to accidental over-compression or unpredictable deformations. The sensors used in the soft robots should be light in weight when compared to the mass of the robot so that the robot performs the intended task with ease. The soft materials used in the soft robot is prone to wear and tear with passage of time. This will reduce the reliability and performance of the interaction with humans. Hence it should be ensured that the functioning of the soft robot is not affected due to this material degradation. One of the key challenge of any robot is to carry  payload. Due to HRI with soft robots, lifting heavy payloads is a challenge. In scenarios such as industries and other structured environments, the HRI for rigid robots excel soft robots. But in the field of medicine \cite{burgner2015continuum}, HRI for robots are needed and performed at a finer level. Since the HRI for soft robots is in the infantry stage, the  technical complexity and cost of production are at the higher end.    

\subsection{Research Opportunities}
\subsubsection{Control}
In the field of modeling and simulation, there is an opportunity to develop models for HRI of soft robots that are computationally efficient and accurate. It should also consider the soft robot dynamics including the material properties, interaction with the environment and non-linearities. The deformation and interaction forces must be sensed using light weight stretchable or flexible sensors embedded on the body of soft robots. Also the efforts must be taken to integrate multi-modal sensing in order to improve perception and situational awareness during the soft robot interaction. The control algorithm should be capable of handling uncertainties and variations (such as environmental conditions, material properties, noise in the sensor data) during interaction with soft robot. The field of learning based control can be utilized since the soft robots can learn from the interaction experience and optimize their control policies over a period of time. The property of compliance and deformation can be studied in detail and can be used in HRI with soft robots. These control strategies should also prioritize safety during HRI. There is a good possibility of creating more intractable and responsive soft robotic systems if there is innovation and development in soft material design and actuation technologies. 

\subsubsection{Design}
There are a lot of opportunities in the field of design for HRI of soft robotics. If the field of materials is explored using tailored mechanical properties such as stiffness, compliance and stretchability, the capabilities of the soft robots can be increased. There are challenges in integrating multiple materials to fabricate a soft robot. The developments in the field of adhesion can lead to better integration of various materials. The design of soft actuators and soft sensors that are precise in rough environments will lead to a robust system that can interact with humans. Although there are advances in the field of manufacturing in the past 2 decades, there is still scope for development of manufacturing procedures that can use multiple materials to fabricate a single soft robot. By exploring the design of soft robots, researchers can identify and improve key requirements for HRI in soft robots, such as safety and compliance, ergonomics, sensory feedback, social interaction, and the ability to collaboratively work with humans.     

\subsubsection{Materials}
The development of novel elastomers and polymers that have tunable mechanical properties can aid the development of robust soft robots. The usages of stimuli-responsive materials such as shape-memory polymers can be used to create innovative HRI for soft robots. The materials that are flexible and conductive in nature can be used to fabricate the soft sensors that are present in areas that bend. The usage of bio-compatible materials for building soft robots find applications in the field of medicine and wearable devices. In the recent years, there are self healing polymers that are available, which when damaged, repairs itself \cite{terryn2021review}. These self healing polymers can prolong the life of the soft robot even if it is damaged during usage. There is a need to study how HRI can be conducted on the soft robots which are made of materials that can withstand harsh environments such as extreme temperatures and chemical exposures. The material science experts should collaborate with experts in control systems and robotics in order to develop materials that can be used for enhanced performance and capabilities of soft robots.

\subsubsection{Manufacturing}
The field of manufacturing has matured, but there is a need for high resolution printing techniques for complex soft robot geometries. The field of soft lithography can be investigated for producing soft robot components with precise shape and features. Also  reusable molds and templates need to be developed for achieving production grade soft robots. The merits of various fabrication techniques can be studied and hybrid fabrication techniques can be used to manufacture the soft robot.   

\subsubsection{Modalities}
\paragraph{Input modalities:}
A few of the innovative input modalities that exist in the literature are discussed as follows. The usage of mid-air gestures for interaction is a novel modality. An example where it is applied to the fully autonomous vehicles is shown in \cite{fink2023autonomous, hafizi2023vehicle}. In the recent works \cite{brooks2023smell}, smell is used as an interaction medium. The specially-abled people also would interact with the robot. In the works of \cite{bilius2023understanding}, a study was conducted with wheelchair users to understand with which portion of their body they prefer to interact. It was highlighted that there is a need for personalized gesture sets tailored for the users preference. In the works of \cite{sun2021teethtap}, the teeth was used as an input modality. A total of 13 gestures can be made using the teeth which can be recorded by the wearable earpiece designed by the authors. The face mask was used as an input modality in \cite{yamamoto2023masktrap}, where 12 gestures were identified by the system. The human tongue was used to give 8 input gestures in the works of \cite{gemicioglu2022tongue}. These are a few input modalities proposed in the recent years     
\paragraph{Output modalities:}
The following are the output modalities that can be used in the HRI for soft robots. The emotional perception of facial expressions as done in \cite{herdel2021drone} can be studied and utilized in soft robots. In the works of \cite{Park21}, the authors argue that robots should blush, as it is the ability to display emotions. It can enhance HRI and improve the overall acceptance of social robots. 

Thus these input and output modalities proposed in the recent times can be used for HRI in soft robots.

\section{Conclusions}
\label{sec_conclusion}
In this work, we explored the HRI for soft robots, focusing on control, design, materials, manufacturing, modalities and applications in detail. Based on these details, we discussed the limitation and research opportunities for this field. It was found that the field of controls for soft robotics is difficult and still has open challenges. The field of design has to ensure that the HRI is safe for both the humans and the soft robot. The materials used for the development of soft robots should be wear resistant in order to sustain for a longer time. The manufacturing technology should aid in the easy job, batch and mass production of the interaction device and soft robot. The modalities chosen for interaction should consider humans from varied background. Thus the field of human-robot interaction for soft robots is an emerging field and has a huge potential to grow in the next decade. The article highlights the significant potential for growth and development in HRI for soft robots, by emphasizing that there are huge research opportunities for multimodal interaction with the soft robots during situational impairments of the user.

%\section*{References}

\bibliography{mybibfile}

\newpage
\section{Appendix}

\subsection{Algorithm to find the duplicate files in the databases}
\label{append_duplicate}
\begin{algorithm}
\caption{Find Duplicates in CSV}
\label{algo:find_duplicates}
\begin{algorithmic}[1]
\REQUIRE $files$: List of file paths
\ENSURE $duplicates$: Set of duplicate entries, $duplicates\_count$: Count of duplicate entries

\STATE Initialize an empty list $data\_sets$.
\FOR{each file in $files$}
    \STATE Read the CSV file.
    \STATE Extract data from the first column and store it in a set.
    \STATE Add the set of data to $data\_sets$.
\ENDFOR
\STATE Compute the intersection of all sets in $data\_sets$ and store the result in $duplicates$.
\STATE Compute the count of duplicates ($duplicates\_count$).
\RETURN $duplicates$, $duplicates\_count$.
\end{algorithmic}
\end{algorithm}

\textbf{Main program:}
\begin{algorithmic}[1]
\STATE Define the list of file paths ($files$).
\STATE Call the \textsc{FindDuplicatesInCSV} algorithm with $files$ as input.
\STATE Retrieve the set of duplicate entries and its count.
\IF{$duplicates$ is not empty}
    \STATE Print the number of duplicates found.
    \FOR{each duplicate entry}
        \STATE Print it.
    \ENDFOR
\ELSE
    \STATE Print a message indicating no duplicates are found.
\ENDIF
\end{algorithmic}

\end{document}